\documentclass[11pt]{article}
\usepackage{miao}
\usepackage{mathpazo}
\usepackage[most]{tcolorbox}
\usepackage{authblk}

\usepackage{xcolor}
\usepackage{fancyhdr}
\usepackage[T1]{fontenc}
\usepackage{ltablex} 
\keepXColumns         
\definecolor{myorange}{RGB}{252,129,59}

\usepackage{graphicx} 

\title{\raisebox{-0.37\height}{\includegraphics[height=3.5em]
{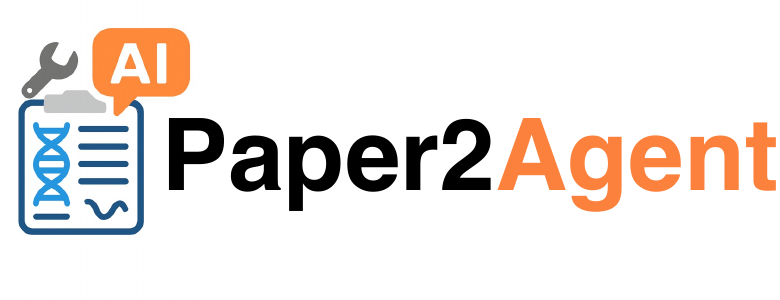}}\,: \textbf{Reimagining Research Papers As Interactive and Reliable AI Agents} }

\author[1,2]{Jiacheng Miao}
\author[1]{Joe R. Davis}
\author[3]{Yaohui Zhang}
\author[1,4]{Jonathan K. Pritchard}
\author[2,3,5]{James Zou}

\affil[1]{Department of Genetics, Stanford University}
\affil[2]{Department of Biomedical Data Science, Stanford University}
\affil[3]{Department of Electrical Engineering, Stanford University}
\affil[4]{Department of Biology, Stanford University}
\affil[5]{Department of Computer Science, Stanford University}

\date{}
\begin{document}
\maketitle
\vspace{-10mm}
\begingroup
\renewcommand\thefootnote{}\footnotetext{\footnotesize	 Email: \{jcmiao, jamesz\}@stanford.edu}%
\addtocounter{footnote}{-1}%
\endgroup
\begin{center}
\href{https://github.com/jmiao24/Paper2Agent}{%
  \raisebox{-0.1\height}{\includegraphics[height=1em]{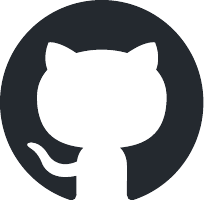}}%
  \hspace{0.3em}\textcolor{myorange}{Repository}
}%
  \quad \quad \quad
\href{https://huggingface.co/spaces/Paper2Agent/alphagenome_agent}{
  \raisebox{-0.25\height}
  {\includegraphics[height=1.5em]{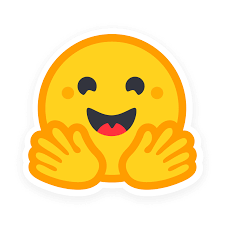}}
  \hspace{0.1em}\textcolor{myorange}{Demo}%
}
\end{center}
\begin{abstract}
{\normalsize
We introduce Paper2Agent, an automated framework that converts research papers into AI agents. Paper2Agent transforms research output from \emph{passive artifacts} into \emph{active systems} that can accelerate downstream use, adoption, and discovery. 
Conventional research papers require readers to invest substantial effort to understand and adapt a paper’s code, data, and methods to their own work, creating barriers to dissemination and reuse. Paper2Agent addresses this challenge by automatically converting a paper into an AI agent that acts as a knowledgeable research assistant. It systematically analyzes the paper and the associated codebase using multiple agents to construct a Model Context Protocol (MCP) server, then iteratively generates and runs tests to refine and robustify the resulting MCP. These paper MCPs can then be flexibly connected to a chat agent (e.g. Claude Code) to carry out complex scientific queries through natural language while invoking tools and workflows from the original paper. We demonstrate Paper2Agent's effectiveness in creating reliable and capable paper agents through in-depth case studies. Paper2Agent created an agent that leverages AlphaGenome to interpret genomic variants and agents based on ScanPy and TISSUE to carry out single-cell and spatial transcriptomics analyses. We validate that these paper agents can reproduce the original paper's results and can correctly carry out novel user queries. Paper2Agent automatically created AI co-scientist that identified new splicing variant associated with ADHD risk. By turning static papers into dynamic, interactive AI agents, Paper2Agent introduces a new paradigm for knowledge dissemination and a foundation for the collaborative ecosystem of AI co-scientists.
}
\end{abstract}

\section{Introduction}

 The research paper is the traditional unit of scientific communication. It remains the norm for documenting methods, results, and insights, and is the primary way research is shared with the broader community. However, papers are fundamentally passive objects: a reader must discover the paper (not an easy task given the flood of publications), parse its contributions, and manually determine how to apply them to their own work. In particular, when a paper describes a new computational method, significant technical barriers often remain before the method can be used on new data \citep{trisovic2022large}. A reader might need to locate the corresponding code repository, install dependencies, configure environments, and interpret the correct inputs and outputs \citep{gomes2022don}. Even with well-maintained repositories, this process is often non-trivial.

For instance, consider AlphaGenome, which provides a powerful framework for genome-scale foundation modeling \citep{avsec2025alphagenome}. Despite its utility, this system requires substantial technical expertise to set up and deploy, limiting accessibility for biologists who could otherwise benefit. Using AlphaGenome in code involves installing the environment, importing multiple modules, creating client objects with API keys, and constructing inputs such as chromosomes, variant objects, and selecting desired output modalities. Users must understand the API hierarchy and parameter semantics, which imposes a learning curve for biologists unfamiliar with these abstractions.

This illustrates a broader challenge: research outputs are passively siloed behind technical barriers. \textbf{Paper2Agent re-imagines research dissemination by turning static papers into active AI agents. Each agent serves as an interactive expert on the corresponding paper, capable of demonstrating, applying, and adapting its methods to new projects.}

\begin{figure}[H]
    \centering
    \includegraphics[width = 0.82\linewidth]{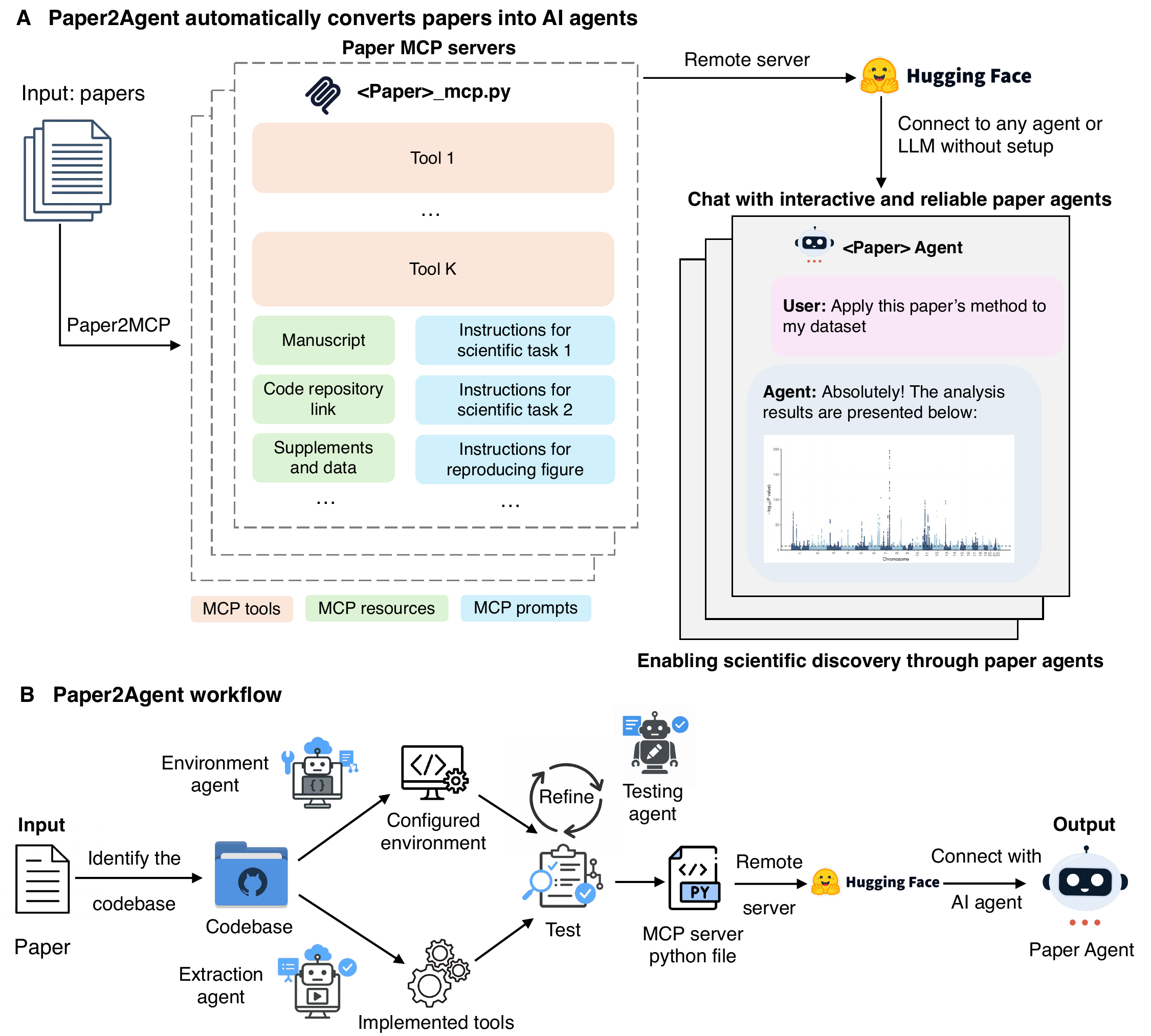}
    \vspace*{-2mm}
        \caption{\small\textbf{Overview of the Paper2Agent.}
        (A) Paper2Agent turns research papers into interactive AI agents by building remote MCP servers with tools, resources, and prompts. Connecting an AI agent to the server creates a paper-specific agent for diverse tasks. (B) Workflow of Paper2Agent. It starts with codebase extraction and automated environment setup for reproducibility. Core analytical features are wrapped as MCP tools, then validated through iterative testing. The resulting MCP server is deployed remotely and integrated with an AI agent, enabling natural-language interaction with the paper’s methods and analyses.}
    \label{fig:1}
\end{figure}

AI agents are autonomous systems that can reason about tasks and act to achieve goals by leveraging external tools and resources \citep{yao2023react}. Modern AI agents are typically powered by large language models (LLMs) connected to external tools or APIs. They can perform reasoning, invoke specialized models, and adapt based on feedback \citep{xia2024agentless}. Agents differ from static models in that they are interactive and adaptive. Rather than returning fixed outputs, they can take multi-step actions, integrate context, and support iterative human–AI collaboration. Importantly, because agents are built on top of LLMs, users can interact with agents through human language, substantially reducing usage barriers for scientists. 

Recent advances highlight the promise of agents for accelerating discovery. For example, the Virtual Lab framework organizes teams of AI scientist agents that collaboratively design and execute research projects across biology and chemistry \citep{swanson2025virtual}. Similarly, Google’s AI co-scientist serves as a virtual collaborator, assisting with hypothesis generation and research proposal development \citep{gottweis2025towards}. Sakana AI’s co-scientist aims for automation of the research lifecycle—from ideation to publication \citep{Sakana.AI2024}. FutureHouse provides an AI scientist platform designed for diverse scientific tasks \citep{ghareeb2025robin}. Alongside these general-purpose platforms, specialized agents are also emerging for specific domains \citep{qu2025crispr}. For example, CellVoyager introduces an agentic system for autonomous analysis of single-cell omics data \citep{alber2025cellvoyager}. Biomni is an AI agent for diverse biological tasks \citep{huang2025biomni}. These systems demonstrate that agents can not only execute code, but also generate hypotheses, evaluate uncertainty, and adapt methods to new datasets. Recent work has also explored automatic code generation from scientific text \citep{seo2025paper2code, movassaghi2025articles}. Paper2Agent complements this emerging paradigm by generalizing the concept: any research paper can be converted into an agent that embodies the knowledge and methods described in the publication.

Paper2Agent provides an automated workflow for converting a scientific paper into an agent. The core idea is to represent the paper as a Model Context Protocol (MCP) server \citep{hou2025model}. MCP is a standardized protocol that allows structured APIs and tools to be exposed in a way that is directly accessible to LLMs and agent frameworks. The conversion process involves: (i) identifying the key contributions of the paper (datasets, methods, models, or workflows); (ii) encapsulating these contributions through an MCP server, defining the inputs, outputs, and usage instructions; (iii) linking the MCP server to LLM-based agents, enabling natural language querying and autonomous execution. Users can then interact with the paper by asking questions, requesting demonstrations, or applying the method to new data.

As an illustration, applying Paper2Agent to AlphaGenome would expose its genome foundation model as an MCP. Instead of requiring users to clone repositories and configure dependencies, they could simply ask: “\textit{Generate AlphaGenome predictions for these variants.}”, "\textit{Interpret the expected effect of this variant on chromatin accessibility in muscle cells.}" or “\textit{Visually compare the AlphaGenome predicted expression changes for a splicing variant in cell types of interest.}” The Paper2Agent-generated agent would handle the setup, execution, and presentation of results, making the method accessible to both computational experts and experimental biologists.

Efforts to make research outputs more executable and accessible have been ongoing for years. Executable papers—such as those proposed in Elsevier’s Executable Paper Grand Challenge \citep{nowakowski2011collage} and more recent Jupyter Notebook–backed publications \citep{rule2019ten}—sought to merge narrative text with runnable code. These approaches increased reproducibility but still required substantial technical familiarity to engage with fully. The Papers with Code initiative \citep{stojnic2019paperswithcode} similarly aimed to bridge papers and implementations by linking publications to open-source repositories. While this improved discoverability, the barrier of installing and executing the code remained.

 Paper2Agent substantially extends this trajectory by providing a new framework: a paper can  be transformed into a capable agent accessible via natural language. In contrast to previous efforts, Paper2Agent shifts the research output from a document or codebase encoding knowledge to a knowledgeable entity capable of execution and dialogue. This represents a new mode of scientific communication, moving beyond static dissemination to interactive collaboration.
 This framework lowers barriers to adoption, democratizes access to advanced methods, and accelerates the translation of research into practice. 

\section{Results}
\subsection*{Overview of Paper2Agent}
Paper2Agent is a multi-agent AI system that automatically transforms research papers into interactive AI agents with minimal human input. The paper agents created via this framework are:
\begin{enumerate}
    \item \textbf{Interactive and easy to use}: Users can execute complex scientific analyses through natural language prompts, eliminating the need for programming expertise.
    \item \textbf{Reliable and reproducible}: Each tool used by a paper agent is validated against the reference codebase’s reported results and figures using example datasets, then locked to ensure reproducibility. This design mitigates the risk of “code hallucination”, where executing inaccurate LLM-generated code could lead to incorrect scientific results. It also minimizes randomness in code generation, further strengthening reproducibility. Finally, every tool includes a code reference from the original paper to provide transparency and traceability.
\end{enumerate}

MCP has recently become an industry standard for connecting LLM-based agents with external resources, providing a unified interface for accessing datasets and tools without custom integration \citep{hou2025model}. Paper2Agent builds on this ecosystem with two components: (i) Paper2MCP, which extracts information from papers and their codebases to build remote MCP servers; and (ii) an agent layer, which wraps each MCP server as a context provider to instantiate paper-specific AI agents (Figure~\ref{fig:1}A). Any LLM or external agent can invoke the servers’ tools through MCP without extra setup. For presentation clarity, we assign one MCP server and one paper agent to each paper. The same approach can create MCPs and agents for a group of related papers. Each MCP server includes three core components:
\begin{enumerate}
    \item \textbf{MCP Tools} are executable functions that encapsulate a paper’s methodological contributions. For example, one AlphaGenome MCP tool takes a genetic variant as input and generates predictions and visualizations of its effects on gene expression, chromatin accessibility, and other modalities. These tools come with a pre-configured environment for seamless execution.
    \item \textbf{MCP Resources} serve as a repository of static assets, including the manuscript text, the associated codebase, and supplementary materials such as datasets, tables, and figures. As an illustration, the AlphaGenome MCP resources include links to the training data used to train the model. All resources are stored in accessible, standardized formats to enable efficient querying and integration by AI agents.
    \item \textbf{MCP Prompts} contain concise instructions that guide AI agents through complex, multi-step scientific workflows derived from a paper’s text or codebase. For example, a Scanpy MCP Prompt encodes the sequence of steps for preprocessing and clustering single-cell data, which we present later in the manuscript. These templates orchestrate tools and resources to ensure reproducible, systematic analyses while reducing the barrier to effective prompting.
\end{enumerate}

The paper MCP servers can be hosted remotely on platforms like 
Hugging Face Spaces, eliminating local dependency issues. MCP standardizes communication, enabling secure and scalable integration with AI agents. The agent layer wraps each Paper2MCP server as a context provider, creating paper-specific conversational agents. Any compatible LLM or agent can connect to these servers to perform tasks such as reproducibility checks, new data analyses, or figure regeneration. For example, a user might ask, “\textit{Apply the method in this paper to the newly generated dataset}”, and the agent will automatically run the pipeline, produce results, and present interpretable outputs. By abstracting away technical details, the agent lowers barriers to method adoption, ensures reproducibility, and helps researchers focus on insights rather than implementation.

We implemented Paper2Agent with Claude Code \citep{anthropic_claude_code}, an AI coding agent specialized in managing complex coding tasks and real-time iterative debugging (Extended Methods). The workflow begins by identifying the codebase associated with a paper (Figure~\ref{fig:1}B). Two specialized agents are then invoked: the environment agent, which configures the necessary software environment, and the extraction agent, which translates core methods into implemented tools. These tools are validated through a testing agent that runs automated checks, refining both the code and environment until results match the reference outputs. Once validated, the tools and environment are packaged into an MCP Python file that can be deployed on a remote server such as Hugging Face. Finally, the paper MCP server is connected with an AI agent to create a fully functional Paper Agent, enabling interactive access to the paper’s methods through natural language queries. We use Claude Code as the downstream AI agent in our case studies, though the paper MCPs can be flexibly integrated with different chat agents. Because MCPs are modular, multiple MCPs can be connected to the same chat agent, enabling users to leverage tools and resources across multiple papers simultaneously.

Next, we present three case studies demonstrating Paper2Agent’s ability to convert diverse research papers into reliable, interactive AI agents for different scientific tasks. These case studies include AlphaGenome \citep{avsec2025alphagenome} for genomics, TISSUE \citep{sun2024tissue} for spatial transcriptomics, and Scanpy \citep{wolf2018scanpy} for single-cell analysis.

\subsection*{AlphaGenome Agent for Genomic Data Interpretation}
The first case study showcases AlphaGenome agent. AlphaGenome is an AI model designed to predict the impact of single-nucleotide variants or mutations in human DNA sequences on a wide range of regulatory processes. Paper2Agent transforms the AlphaGenome paper into an interactive AlphaGenome agent, enabling automated interpretation of genomics data. Through natural language queries, users can leverage this agent to prioritize causal genes for disease-associated variants, clarify the regulatory impact of individual variants, and inform the design of synthetic DNA with specific regulatory functions.

\begin{figure}[H]
    \centering
    \includegraphics[width = 0.98\linewidth]{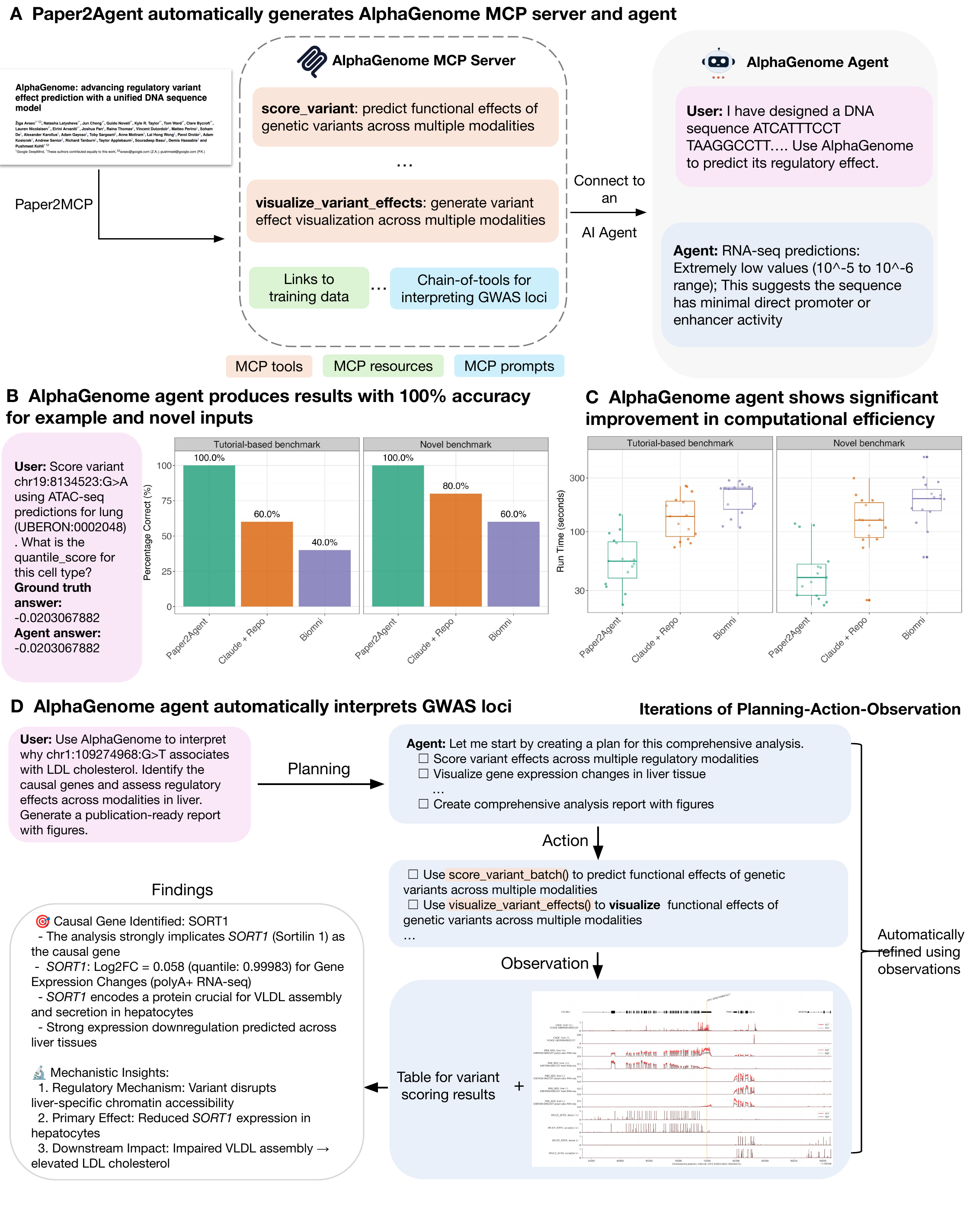}
    \vspace*{-2mm}
        \caption{\textbf{Overview of the Paper2Agent-generated AlphaGenome agent.} (A) Construction of the AlphaGenome MCP server and agent. (B) Benchmarking the AlphaGenome agent on tutorial-based and novel queries shows 100\% accuracy, above that achieved by the general agents Claude with the AlphaGenome codebase (Claude + Repo) or Biomni. (C) The AlphaGenome agent also shows improved computational efficiency in terms of reduced run-time for each query compared to Claude + Repo and Biomni agents. (D) Automated planning and interpretation of GWAS loci through iterative planning–action–observation cycles by the AlphaGenome agent.}
    \label{fig:2}
\end{figure}

Paper2Agent generated 22 AlphaGenome MCP tools in around 3 hours on a personal laptop without human intervention, comprehensively covering its methodological innovations. This one-time process produces reusable tools for future applications. These MCP tools span single- and batch-variant scoring across functional assays, sequence-level prediction, tissue ontology exploration, and an extensive visualization suite (Figure~\ref{fig:2}A). For example, \texttt{score\_variant\_effect()} is an MCP tool that predicts the functional consequences of genetic variants across multiple modalities—such as gene expression, splicing, and chromatin accessibility—within a wide range of tissues and cell types. Complementing this, \texttt{visualize\_variant\_effects()} generates modality-specific visualizations that simplify the interpretation of regulatory impact.

\newpage

Importantly, the tools generated by Paper2Agent are designed with flexible, well-annotated input parameters. For example, the \texttt{visualize\_variant\_effects()} tool exposes a rich set of options that make it adaptable to diverse use cases (Supplementary Figure 1). Given an input genetic variant, the AlphaGenome agent can select the organism to analyze (human or mouse), adjust the sequence context length around the variant,  toggle different modalities—such as RNA-seq, ATAC-seq,  or ChIP-seq histone tracks. Moreover, each MCP tool embeds a traceable link to the original GitHub source code, ensuring transparency and reproducibility. By connecting an AI agent with the AlphaGenome MCP, the system creates the AlphaGenome agent.

Next, we benchmarked the Paper2Agent-generated AlphaGenome agent in producing numerical and qualitative results and figures relative to human experts configuring and running the code manually (Figure ~\ref{fig:2}B). We also compared the AlphaGenome agent to two other agentic systems: 1) Claude Code with access to the AlphaGenome repo (referred to as Claude + Repo) and 2) Biomni. We manually curated 15 example queries directly from the AlphaGenome tutorial, such as "\textit{Score variant chr3:58394738:A>T using ATAC-seq predictions for motor neuron cells (CL:0000100). What is the quantile\_score for this cell type?}", "\textit{Make DNase-seq predictions for sequence 'GATTACA' (padded to 2048 length) for lung tissue (UBERON:0002048). What is the nonzero\_mean value in the dnase metadata}". The AlphaGenome agent achieved 100.0\% (15/15) accuracy on these queries, higher than Claude + Repo at 60.0\% (9/15), and Biomni at 40.0\% (6/15). 

To assess generalizability and guard against potential overfitting to the original examples, we also manually curated a set of novel queries that were not present in either the paper or its codebase. These included previously untested variant positions, allelic substitutions, and tissue–cell type contexts, such as "\textit{Analyze variant chr9:98765432:T>C with DNASE predictions for muscle cells (CL:0000187). What is the quantile\_score for muscle tissue?" and "Analyze histone ChIP-seq metadata for neuronal stem cells. What is the nonzero\_mean value for H3K4me3 in neuronal stem cells (CL:0000100)}?" The AlphaGenome agent again achieved 100.0\% (15/15) accuracy, as verified by manual execution of the original AlphaGenome code. Claude + Repo achieved the next highest accuracy at 80.0\% (12/15) and Biomni solved 60.0\% (9/15) of the queries correctly. 

The AlphaGenome agent also consistently outperformed both agent systems in terms of compute efficiency, showing a median decrease in run time for the tutorial-based benchmark of 1.8x and 3.1x compared to Claude + Repo and Biomni, respectively (Figure ~\ref{fig:2}C). For the novel benchmark, we observed a median run time improvement of 3.2x and 4.6x, respectively. These results highlight the improved reliability and efficiency achieved by Paper2Agent-generated agents compared to other agentic systems.

Finally, we demonstrated that the AlphaGenome agent enables automatic interpretation of Genome-Wide Association Study (GWAS) loci and validation of the analysis in the original paper. We considered the example of interpreting why the genetic variant chr1:109274968:G>T is associated with low-density lipoprotein cholesterol that was presented in the original AlphaGenome paper (Figure ~\ref{fig:2}D). Based on the tools available, the AlphaGenome agent constructs a step-by-step plan to solve this task. This plan includes generating input files, scoring variants across multiple modalities, filtering results for trait-relevant tissues, creating modality-specific visualizations (chromatin accessibility, histone marks, transcription factor binding, and splicing), and assembling a comprehensive interpretation report. The agent then executes these actions using implemented tools, such as \texttt{score\_variant()} and \texttt{visualize\_tf\_binding()}, automatically refining its strategy through iterative observation and feedback. A final report is then presented to provide a unified interpretation of the regulatory impact of the variant, integrating evidence across modalities and tissues.

Interestingly, the AlphaGenome agent prioritizes \textit{SORT1} as the most likely causal gene, whereas the original paper emphasized \textit{CELSR2} and \textit{PSRC1}. The agent favors \textit{SORT1} for two reasons: 1) a high quantile score (0.99982) indicating a strong predicted impact on \textit{SORT1} expression in liver tissue. Here, the quantile score reflects how extreme the variant’s predicted effect relative to other variants 2) \textit{SORT1} encodes sortilin, directly involved in LDL/VLDL secretion \citep{kjolby2015sortilin}. We manually queried the GTEx \citep{gtex2020gtex} eQTL data and confirmed that this variant is a significant eQTL for \textit{SORT1} (p = 1.1e-65) in liver. However, both \textit{CELSR2} and \textit{PSRC1} also exhibit high AlphaGenome quantile scores (0.99998 each) and significant eQTL associations in GTEx liver (p =
4.7e-46 and 8.5e-50, respectively).
This result shows the inherent difficulty in confidently assigning causal genes at complex GWAS loci where the variants are eQTLs for multiple nearby genes \citep{wainberg2019opportunities, mostafavi2023systematic}.

This discrepancy highlights a key strength of Paper2Agent: with a single prompt, users can re-evaluate published conclusions using independent model-based evidence. Rather than treating the original interpretation as fixed, the agent enables dynamic hypothesis re-assessment and, at scale, provides a systematic way to revisit conclusions across many studies.

\subsection*{TISSUE Agent for Uncertainty-Aware Single-Cell Spatial Transcriptomics Analysis}

\begin{figure}[H]
    \centering
    \includegraphics[width = 0.9\linewidth]{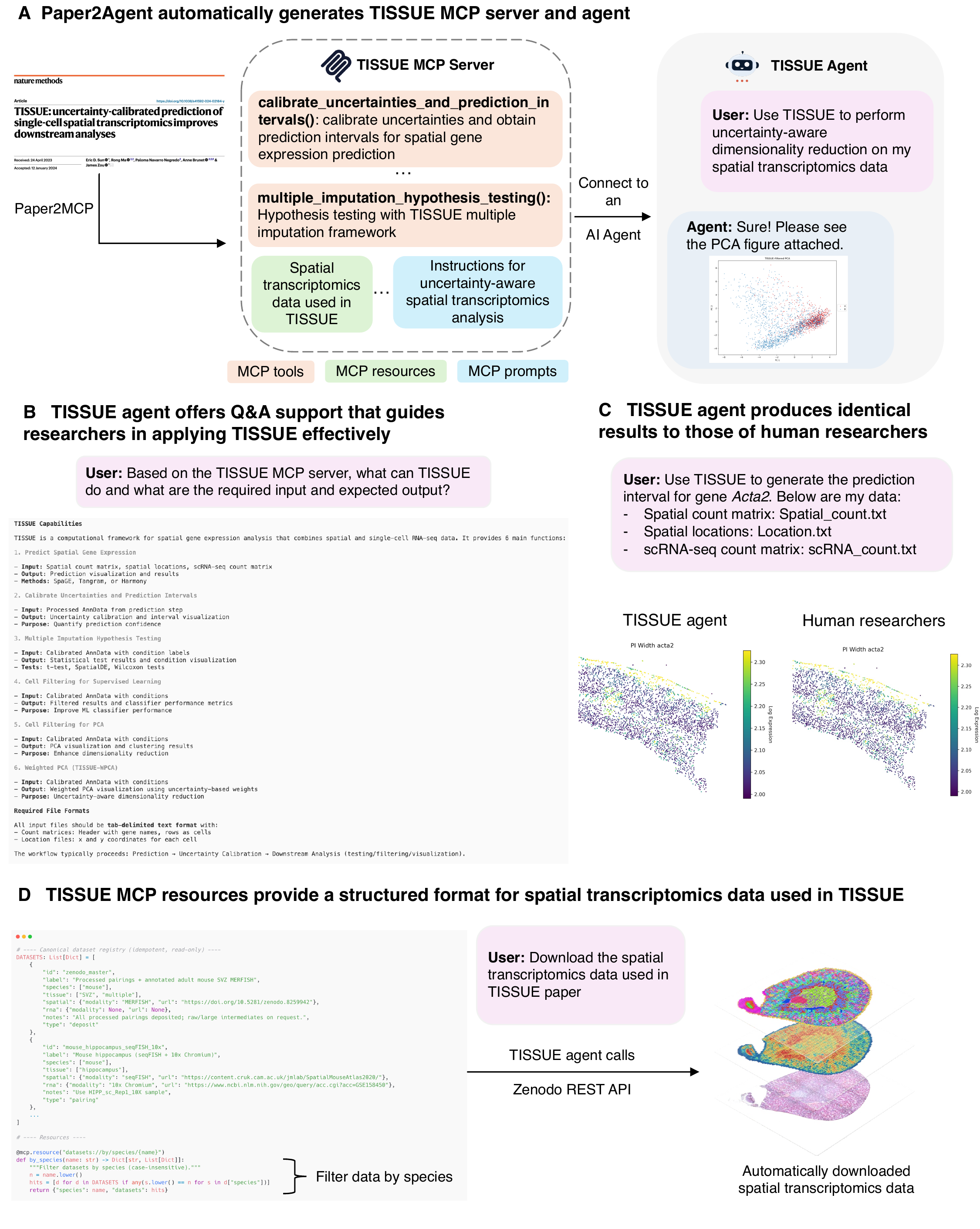}
    \vspace*{-2mm}
        \caption{\small\textbf{Overview of the Paper2Agent-generated TISSUE agent.} (A) Construction of the TISSUE MCP server and agent. (B) Q\&A support for uncertainty-aware spatial transcriptomics analysis. (C) Reproducibility confirmed by matching human researcher results. (D) Structured MCP resources enable standardized dataset access and automated downloads.}
    \label{fig:3}
\end{figure}

We next present the Paper2Agent-generated agent for TISSUE \citep{sun2024tissue}, a recent paper that developed a new method for uncertainty-aware single-cell spatial transcriptomics analysis (Figure~\ref{fig:3}A). This case study reflects a common scenario: a new methodology paper is published, and researchers want to apply the method to their own data but lack the time to navigate the codebase, configure the environment, and grasp the method’s features and input requirements. Paper2Agent addresses these challenges by automatically generating ready-to-use agents for diverse papers and providing Q\&A support to guide input preparation and clarify what the method can do.

Paper2Agent generated 6 tools for the TISSUE MCP server, covering spatial gene expression prediction, prediction interval construction, and uncertainty-aware downstream analysis such as hypothesis testing, prediction, and dimensionality reduction (Figure~\ref{fig:3}A). Importantly, the TISSUE agent can also serve as an interactive guide (Figure~\ref{fig:3}B). For example, when prompted with “\textit{Based on the TISSUE MCP server, what are the required inputs for TISSUE?}”, the agent returns a structured and comprehensive explanation of the method’s required inputs, expected outputs, and available features. This transforms the TISSUE paper into an interactive AI agent: instead of manually searching through documentation or code, users can directly ask the agent about how to use TISSUE and receive precise, actionable instructions.

Next, we evaluate the TISSUE agent’s ability to construct prediction intervals for spatial transcriptomic (ST) prediction. We prompt the agent "C\textit{alculate the prediction interval for the spatial gene expression prediction of
gene Acta2 using TISSUE. This is my data: Spatial count matrix: Spatial\_count.txt Spatial locations:
Locations.txt scRNA-seq count matrix: scRNA\_count.txt}". The agent automatically executes the TISSUE pipeline, without additional user intervention (Figure~\ref{fig:3}C). The output matches the results obtained by human experts running the pipeline manually. This illustrates the paper agent's ability to run entire analysis workflows (in this case, from data loading and preprocessing through imputation and uncertainty estimation), not just individual tools.

Finally, we showcase the use of MCP resources by translating the data availability section of the TISSUE paper into a structured registry. This registry harmonizes ST datasets with standardized metadata (species, tissue type, modality, and data URL) and makes them directly accessible to the TISSUE agent through data repository APIs such as the Zenodo REST API (Figure~\ref{fig:3}D). Users can query and filter datasets, for example, by species—without manually navigating multiple repositories. Combined with the TISSUE MCP tools, a user's query might be: “\textit{Download the mouse spatial transcriptomics data from this paper and run TISSUE to generate a prediction interval after applying spatial prediction to the dataset.}”. The TISSUE agent then automatically filters for mouse data, downloads it, and applies the TISSUE pipeline.

\begin{figure}[H]
    \centering
    \includegraphics[width = 1\linewidth]{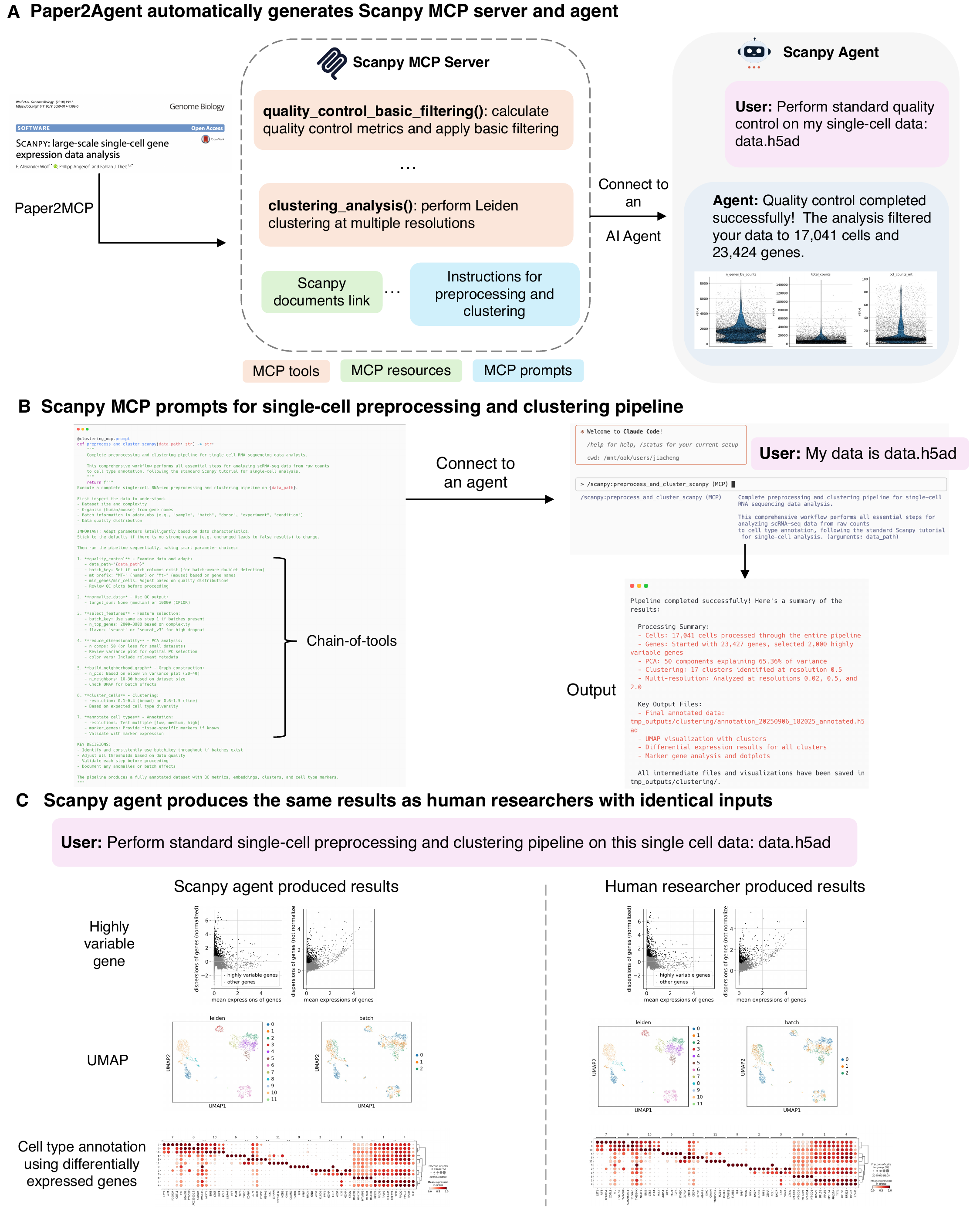}
    \vspace*{-2mm}
        \caption{\textbf{Overview of the Paper2Agent-generated Scanpy agent.} (A) Construction of the Scanpy MCP server and agent. (B) MCP prompts encode a standardized single-cell preprocessing and clustering pipeline. (C) Agent reproduces human researcher results, requiring only the dataset path as input.}
    \label{fig:4}
\end{figure}

\subsection*{Scanpy Agent for Single-Cell Data Preprocessing}
Next, we demonstrate the application of the Paper2Agent-generated Scanpy agent for preprocessing and clustering for single-cell data analysis. Scanpy is a widely used, comprehensive package for analyzing large-scale single-cell transcriptomic data \citep{wolf2018scanpy}. In practice, many workflows in single-cell data analysis may rely on only a subset of Scanpy's features. To accommodate this, Paper2Agent supports converting not only entire methods but also specific parts of a paper’s method into tools, enabling the AI agent to expose only the features most relevant for a given analysis.

We focus on Scanpy’s most common use case: preprocessing and clustering single-cell data. Paper2Agent generates 7 tools for this feature in around 45 minutes on a personal laptop -- tools such as \texttt{quality\_control()} for calculating and visualizing QC metrics, filtering cells and genes, and detecting doublets, and \texttt{normalize\_data()} for normalizing count data (Figure~\ref{fig:4}A). This allows users to prompt the Scanpy agent to perform quality control on their single-cell data.

In practice, many users prefer an end-to-end workflow for preprocessing and clustering, where the implemented tools are executed sequentially in the correct order. This type of analysis workflow is not unique to single-cell analysis but is common across many scientific domains. However, executing such workflows can be challenging: the AI agent must either already “know” the correct order of actions, or the user must provide a carefully structured prompt that explicitly specifies the sequence. To overcome this limitation, we use MCP prompts to guide the agent. MCP prompts offer a standardized way to encode workflows, ensuring that tools are executed in the proper order and relieving users from the burden of manually instructing the agent. Importantly, these MCP prompts are inferred directly from the paper and codebase by Paper2Agent, without the need for manual curation. This design improves both reproducibility and usability, particularly for complex analyses such as single-cell data processing.

For example, the Paper2Agent-generated Scanpy MCP prompts encode a standard preprocessing and clustering pipeline, including quality control, normalization, feature selection, dimensionality reduction, graph construction, clustering, and cell-type annotation in the correct order (Figure~\ref{fig:4}B). The prompt also instructs the Scanpy agent to inspect the data before analysis to select appropriate parameters. Users only need to provide the data path (e.g., data.h5ad), and the Scanpy agent automatically runs the workflow and provides a summary of the analysis results.

To evaluate the Scanpy agent’s performance, we applied it to preprocess and cluster three publicly available single-cell datasets from 10x Genomics (Data availability) that are not included in the Scanpy codebase. We invoke the Scanpy MCP prompts and query the Scanpy agent "\textit{Perform standard single-cell preprocessing and clustering pipeline on this single-cell data: data.h5ad}". As shown in Figure~\ref{fig:4}C, the agent produces outputs that match those produced by human researchers when processing the same data. This demonstrates how MCP-prompt–powered Scanpy agents streamline workflow execution, making advanced single-cell analysis both accessible and reproducible.

\subsection*{Paper2Agent enables autonomous AI-driven collaboration and discovery}

Human scientific collaboration often advances by combining insights from multiple, disparate papers—for instance, when a newly developed method is applied to a recently published dataset to generate fresh discoveries. However, this process is typically slow and labor-intensive. Paper2Agent enables a new mode of AI-driven collaboration in which AI paper agents can interact directly with each other. The agent of a new method paper can autonomously collaborate with the agent of a new data paper to perform analyses, test hypotheses, and generate new insights.


As an illustrative case study, we consider the AlphaGenome method paper \citep{avsec2025alphagenome} and a recent GWAS of Attention-Deficit/Hyperactivity Disorder (ADHD) data and discovery paper \citep{van2025genome}, representing a common scenario where a new method and a new dataset/discovery become available. Researchers may want to collaborate to integrate AlphaGenome with the data and insights from the GWAS study, but doing this manually can take weeks. 

 Paper2Agent automatically creates MCPs for AlphaGenome and the ADHD GWAS data. The data agent in Paper2Agent automatically converts the paper-associated datasets—such as released GWAS summary statistics and supplementary tables reporting genetic discoveries—into standardized MCP resources, enabling seamless integration with analytical agents for downstream analyses. 
 
 These MCPs are then automatically connected with a downstream AI co-scientist. Here we use Claude Code as our agentic system. This Paper2Agent-generated AI co-scientist autonomously generates new hypotheses, and designs and implements analyses to explore each hypothesis. The AI co-scientist proposed several interesting hypotheses to explore, including: 1) ADHD risk variants alter regulatory activity in brain-specific cell types. 2) AlphaGenome prioritizes causal variants within ADHD fine-mapping credible sets. 3) ADHD-associated variants disrupt transcription factor binding at \textit{FOXP} family gene loci. Users can interact with the AI co-scientist in real-time, guiding it to execute the research plan for hypothesis (2) and to explore the mechanisms of the identified causal variant. Combining insights from the AlphaGenome and ADHD GWAS papers, the AI co-scientist prioritizes a single causal variant from 209 candidates identified in the fine-mapping credible sets and elucidates its molecular mechanisms contributing to ADHD risk. The agent showed that the intronic variant rs1626703 is predicted to alter \textit{MPHOSPH9} splicing and expression specifically in glutamatergic neurons (AlphaGenome quantile scores: splice junction = 1.000; RNA-seq = 0.963). AlphaGenome  indicates that rs1626703 promotes exon inclusion, as evidenced by increased read coverage and strengthened splice junctions in the alternate allele, resulting in elevated \textit{MPHOSPH9} expression. \textit{MPHOSPH9} encodes an M-phase–associated phosphoprotein that localizes to centrosomes/centrioles and has been implicated in the recruitment of the CP110-CEP97 complex during cell division and ciliogenesis. \citep{huang2018m}.

Furthermore, the AI co-scientist autonomously uses AlphaGenome to prioritize causal variants across all 39 loci, completing the analysis within two hours. It automatically constructs a systematic workflow that extracts credible-set variants, performs AlphaGenome functional scoring for those variants in glutamatergic neurons, filters for protein-coding genes, and ranks variants by their maximum quantile impact scores across different functional modalities. For each locus, the AI co-scientist identifies the top variant, maps it to its target gene, summarizes its molecular effects, and compiles a comprehensive markdown report detailing the functional consequences and biological significance of the prioritized gene. The AI-generated reports for the causal variants and their mechanisms aross all 39 loci are provided in Supplementary Table 1. This workflow enables scalable hypothesis generation and mechanistic interpretation of GWAS loci in hours rather than weeks of manual review. While these hypotheses benefit from subsequent human evaluation, the workflow shifts scientific effort from manual execution to the synthesis of actionable biological insights.

Together, these results exemplify a new collaborative paradigm in which human scientists use Paper2Agent to design their own AI co-scientists and jointly formulate high-level scientific hypotheses, while the AI co-scientists autonomously execute and interpret complex analytical tasks. 

\begin{figure}[H]
    \centering
    \includegraphics[width = 1\linewidth]{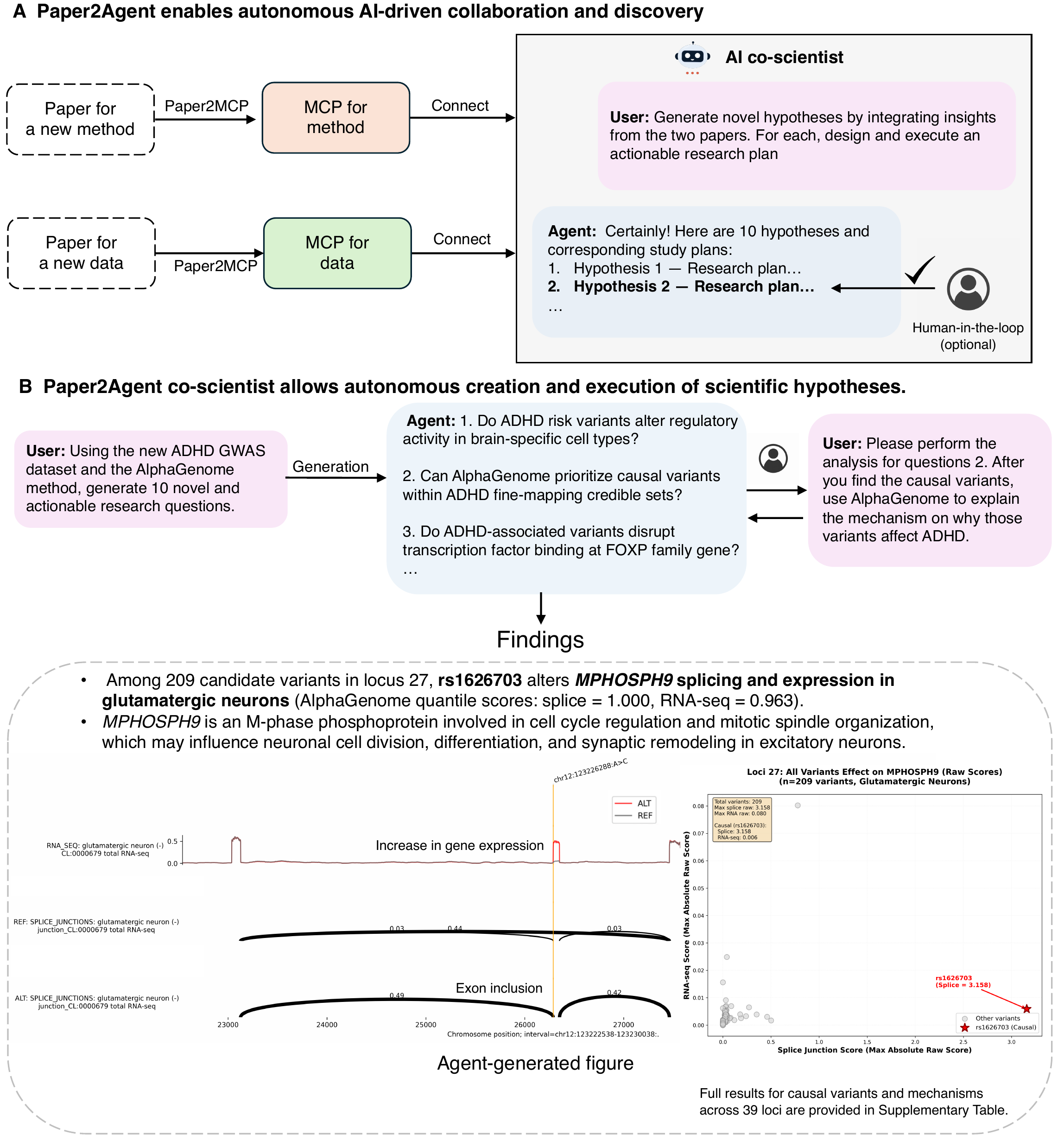}
    \vspace*{-2mm}
        \caption{\textbf{Paper2Agent enables autonomous AI-driven collaboration and genomic discovery.} (A) Paper2Agent transforms scientific papers into Model Context Protocol (MCP) resources for both methods and data, allowing AI co-scientist to integrate them and autonomously generate novel hypotheses and actionable research plans. (B) Using the ADHD GWAS dataset and the AlphaGenome method MCPs, the agent autonomously generates and tests scientific hypotheses, identifies causal variants, and interprets molecular mechanisms. The agent identified rs1626703 as a likely causal variant among 209 candidate variants. The agent then showed computationally using AlphaGenome that rs1626703 alters splicing of \textit{MPHOSPH9} and increases its expression in glutamatergic neurons, revealing a plausible causal mechanism for ADHD risk.}
    \label{fig:5}
\end{figure}

\section{Discussion}
In this work, we introduce Paper2Agent, a framework that transforms a research paper from a passive publication into an interactive AI agent. We demonstrate this approach by creating Paper2Agent instances for several methodological advances, including AlphaGenome for genomics, Scanpy for single-cell analysis, and TISSUE for spatial transcriptomics. These examples illustrate how a paper agent can embody the research contribution, making it directly accessible through natural language interaction. The generated paper MCPs are modular units that can be connected to diverse user-facing agents, enabling broad adoption. By lowering the barrier between publication and practical application, Paper2Agent helps bridge the gap between how scientific discoveries are disseminated and how they are used in practice.

Our initial focus has been on methodological papers, since they offer the clearest use case. Such papers typically describe algorithms, models, or computational workflows that other researchers seek to adopt, but whose deployment often requires substantial technical expertise. Converting them into agents allows the methods to be applied to new problems without the overhead of mastering complex software ecosystems. In future work, we plan to further expand Paper2Agent to other forms of research output, including data resources and discovery papers. In those contexts, the agent’s role may shift from computation to interpretation, curation, or explanation, guiding users through datasets or contextualizing new insights for diverse scientific communities.

Not every paper can be seamlessly turned into a robust agent. If the original codebase is incomplete, poorly documented, or contains unresolved errors, Paper2Agent cannot reliably expose it as a functioning tool. Yet this limitation is itself informative: the ease with which a paper can be transformed into an agent can serve as a practical measure of reproducibility and rigor. Just as the scientific community has come to expect clear data and code availability, we envision that a natural extension will be to expect contributions to be structured in ways that facilitate their translation into agents. Well-documented, modular, and transparent papers will naturally lend themselves to this new standard.

To better quantify this ease of reproducibility and agentification, we used a benchmarking approach based on human expert-evaluated examples from the paper as well as novel examples meant to test generalizability. With this approach, we showed, for example, that the AlphaGenome agent was able to execute both tutorial-based and novel queries with 100.0\% accuracy, much higher than the accuracy achieved by other agents like Claude Code and Biomni. Moreover, the agent created by Paper2Agent is substantially more efficient, 
 achieving median run time speedups of  over 1.7x compared to Biomni or using Claude Code direct on the repo. Our benchmarking approach, however, is limited by expert knowledge of the paper and method and manual implementation and review. A future direction is to further streamline this process with additional agentic frameworks, e.g. with LLM-as-judge evaluations \citep{gu2024survey}.

Another consideration is the scope of agentification. While the paper is the conventional unit of scientific communication, it is not always the best unit for agentification. In many fields, an idea evolves across a sequence of publications, each adding refinements, benchmarks, or applications. In such cases, the most useful agent may not represent a single paper but rather a collection of related works aggregated into a coherent interface. A single MCP can encapsulate multiple related papers. We plan to extend Paper2Agent to flexibly accommodate this broader scope.

Looking forward, just as many journals now require data and code availability sections, we anticipate the emergence of an “agent availability” section that specifies whether and how the contribution has been embodied as an interactive agent. This would not only provide immediate utility to readers but also incentivize authors to present their work in a form conducive to agentification.

Finally, once scientific knowledge is encoded in active agents rather than static artifacts, the potential extends beyond individual use. Agents could interact with one another, linking methods to datasets or combining insights from different domains, as illustrated in our AlphaGenome-ADHD case study. 
Communities of such agents could form a dynamic layer of scientific intelligence, accelerating connections across disciplines and enabling a new form of AI-driven collaboration. Paper2Agent thus points toward a future in which scientific communication is not only about describing results, but also about creating interactive, collaborative entities that embody and extend the research.

\section*{Data availability}
This paper utilized publicly available data for analysis:
\begin{description}{
\normalsize
\item 10x Genomics PBMC single-cell RNA-seq datasets:
\path{http://cf.10xgenomics.com/samples/cell-exp/3.0.0/pbmc_1k_v2/pbmc_1k_v2_filtered_feature_bc_matrix.h5}, 
\path{http://cf.10xgenomics.com/samples/cell-exp/3.0.0/pbmc_1k_v3/pbmc_1k_v3_filtered_feature_bc_matrix.h5}, 
\path{http://cf.10xgenomics.com/samples/cell-exp/3.0.0/pbmc_1k_protein_v3/pbmc_1k_protein_v3_filtered_feature_bc_matrix.h5}.

\item{Mouse somatosensory cortex spatial transcriptomic data:} Dataset15 in \path{https://zenodo.org/records/8259942}

 \item{GTEx portal:} \path{https://gtexportal.org/home/}

 \item{AlphaGenome Github repository: https://github.com/google-deepmind/alphagenome}
 \item{TISSUE Github repository: https://github.com/sunericd/TISSUE}

 \item{Scanpy Github repository: https://github.com/scverse/scanpy}
}
\end{description}

\section*{Code availability}

Paper2Agent is publicly available at \url{https://github.com/jmiao24/Paper2Agent}.\\
AlphaGenome MCP server: \url{https://huggingface.co/spaces/Paper2Agent/alphagenome_mcp}.\\
Scanpy MCP server: \url{https://huggingface.co/spaces/Paper2Agent/scanpy_mcp}.\\
TISSUE MCP server: \url{https://huggingface.co/spaces/Paper2Agent/tissue_mcp}.

\section*{Agent availability}
Paper2Agent-generated AlphaGenome agent is publicly available at \url{https://huggingface.co/spaces/Paper2Agent/alphagenome_agent}.

\section*{Acknowledgments} We thank Abubakar Abid, Eric Sun, Emma Dann, lab members from the Zou lab and the Pritchard lab for helpful feedback during the project. J.Z. is supported by funding from the Chan-Zuckerberg Biohub.

\bibliographystyle{unsrt}
\bibliography{main}

\begin{thebibliography}{10}

\bibitem{trisovic2022large}
Ana Trisovic, Matthew~K. Lau, Thomas Pasquier, and Merc{\`e} Crosas.
\newblock A large-scale study on research code quality and execution.
\newblock {\em Scientific Data}, 9(1):60, 2022.

\bibitem{gomes2022don}
Dylan~G.~E. Gomes, Patrice Pottier, Robert Crystal-Ornelas, Emma~J. Hudgins,
  Vivienne Foroughirad, Luna~L. S{\'a}nchez-Reyes, Rachel Turba, Paula~Andrea
  Martinez, David Moreau, Michael~G. Bertram, et~al.
\newblock Why don't we share data and code? perceived barriers and benefits to
  public archiving practices.
\newblock {\em Proceedings of the Royal Society B}, 289(1987):20221113, 2022.

\bibitem{huang2018m}
Ning Huang, Donghui Zhang, Fangyuan Li, Peiyuan Chai, Song Wang, Junlin Teng,
  and Jianguo Chen.
\newblock M-phase phosphoprotein 9 regulates ciliogenesis by modulating
  cp110-cep97 complex localization at the mother centriole.
\newblock {\em Nature Communications}, 9(1):4511, 2018.

\bibitem{rule2019ten}
Adam Rule, Amanda Birmingham, Cristal Zuniga, Ilkay Altintas, Shih-Cheng Huang,
  Rob Knight, Niema Moshiri, Mai~H. Nguyen, Sara~Brin Rosenthal, Fernando
  P{\'e}rez, et~al.
\newblock Ten simple rules for writing and sharing computational analyses in
  jupyter notebooks.
\newblock {\em PLoS Computational Biology}, 15(7):e1007007, 2019.

\bibitem{kjolby2015sortilin}
Mads Kjolby, Morten~Schallburg Nielsen, and Claus~Munck Petersen.
\newblock Sortilin, encoded by the cardiovascular risk gene sort1, and its
  suggested functions in cardiovascular disease.
\newblock {\em Current Atherosclerosis Reports}, 17(4):18, 2015.

\bibitem{gtex2020gtex}
GTEx Consortium.
\newblock The gtex consortium atlas of genetic regulatory effects across human
  tissues.
\newblock {\em Science}, 369(6509):1318--1330, 2020.

\bibitem{wainberg2019opportunities}
Michael Wainberg, Nasa Sinnott-Armstrong, Nicholas Mancuso, Alvaro~N. Barbeira,
  David~A. Knowles, David Golan, Raili Ermel, Arno Ruusalepp, Thomas
  Quertermous, Ke~Hao, et~al.
\newblock Opportunities and challenges for transcriptome-wide association
  studies.
\newblock {\em Nature Genetics}, 51(4):592--599, 2019.

\bibitem{nowakowski2011collage}
Piotr Nowakowski, Eryk Ciepiela, Daniel Har{\k{e}}{\.z}lak, Joanna Kocot, Marek
  Kasztelnik, Tomasz Barty{\'n}ski, Jan Meizner, Grzegorz Dyk, and Maciej
  Malawski.
\newblock The collage authoring environment.
\newblock {\em Procedia Computer Science}, 4:608--617, 2011.

\bibitem{wolf2018scanpy}
F.~Alexander Wolf, Philipp Angerer, and Fabian~J. Theis.
\newblock Scanpy: large-scale single-cell gene expression data analysis.
\newblock {\em Genome Biology}, 19(1):15, 2018.

\bibitem{sun2024tissue}
Eric~D. Sun, Rong Ma, Paloma Navarro~Negredo, Anne Brunet, and James Zou.
\newblock Tissue: uncertainty-calibrated prediction of single-cell spatial
  transcriptomics improves downstream analyses.
\newblock {\em Nature Methods}, 21(3):444--454, 2024.

\bibitem{mostafavi2023systematic}
Hakhamanesh Mostafavi, Jeffrey~P. Spence, Sahin Naqvi, and Jonathan~K.
  Pritchard.
\newblock Systematic differences in discovery of genetic effects on gene
  expression and complex traits.
\newblock {\em Nature Genetics}, 55(11):1866--1875, 2023.

\bibitem{yao2023react}
Shunyu Yao, Jeffrey Zhao, Dian Yu, Nan Du, Izhak Shafran, Karthik Narasimhan,
  and Yuan Cao.
\newblock React: synergizing reasoning and acting in language models.
\newblock In {\em International Conference on Learning Representations (ICLR)},
  2023.

\bibitem{avsec2025alphagenome}
{{Z}}iga Avsec, Natasha Latysheva, Jun Cheng, Guido Novati, Kyle~R. Taylor, Tom
  Ward, Clare Bycroft, Lauren Nicolaisen, Eirini Arvaniti, Joshua Pan, et~al.
\newblock Alphagenome: advancing regulatory variant effect prediction with a
  unified dna sequence model.
\newblock {\em bioRxiv}, pages 2025--06, 2025.

\bibitem{swanson2025virtual}
Kyle Swanson, Wesley Wu, Nash~L. Bulaong, John~E. Pak, and James Zou.
\newblock The virtual lab of ai agents designs new sars-cov-2 nanobodies.
\newblock {\em Nature}, pages 1--3, 2025.

\bibitem{hou2025model}
Xinyi Hou, Yanjie Zhao, Shenao Wang, and Haoyu Wang.
\newblock Model context protocol (mcp): landscape, security threats, and future
  research directions.
\newblock {\em arXiv preprint arXiv:2503.23278}, 2025.

\bibitem{gottweis2025towards}
Juraj Gottweis, Wei-Hung Weng, Alexander Daryin, Tao Tu, Anil Palepu, Petar
  Sirkovic, Artiom Myaskovsky, Felix Weissenberger, Keran Rong, Ryutaro Tanno,
  et~al.
\newblock Towards an ai co-scientist.
\newblock {\em arXiv preprint arXiv:2502.18864}, 2025.

\bibitem{ghareeb2025robin}
Ali~Essam Ghareeb, Benjamin Chang, Ludovico Mitchener, Angela Yiu, Caralyn~J.
  Szostkiewicz, Jon~M. Laurent, Muhammed~T. Razzak, Andrew~D. White,
  Michaela~M. Hinks, and Samuel~G. Rodriques.
\newblock Robin: a multi-agent system for automating scientific discovery.
\newblock {\em arXiv preprint arXiv:2505.13400}, 2025.

\bibitem{qu2025crispr}
Yuanhao Qu, Kaixuan Huang, Ming Yin, Kanghong Zhan, Dyllan Liu, Di~Yin,
  Henry~C. Cousins, William~A. Johnson, Xiaotong Wang, Mihir Shah, et~al.
\newblock Crispr-gpt for agentic automation of gene-editing experiments.
\newblock {\em Nature Biomedical Engineering}, pages 1--14, 2025.

\bibitem{alber2025cellvoyager}
Samuel Alber, Bowen Chen, Eric Sun, Alina Isakova, Aaron~J. Wilk, and James
  Zou.
\newblock Cellvoyager: ai compbio agent generates new insights by autonomously
  analyzing biological data.
\newblock {\em bioRxiv}, pages 2025--06, 2025.

\bibitem{huang2025biomni}
Kexin Huang, Serena Zhang, Hanchen Wang, Yuanhao Qu, Yingzhou Lu, Yusuf
  Roohani, Ryan Li, Lin Qiu, Gavin Li, Junze Zhang, et~al.
\newblock Biomni: a general-purpose biomedical ai agent.
\newblock {\em bioRxiv}, pages 2025--05, 2025.

\bibitem{xia2024agentless}
Chunqiu~Steven Xia, Yinlin Deng, Soren Dunn, and Lingming Zhang.
\newblock Agentless: demystifying llm-based software engineering agents.
\newblock {\em arXiv preprint arXiv:2407.01489}, 2024.

\bibitem{gu2024survey}
Jiawei Gu, Xuhui Jiang, Zhichao Shi, Hexiang Tan, Xuehao Zhai, Chengjin Xu, Wei
  Li, Yinghan Shen, Shengjie Ma, Honghao Liu, et~al.
\newblock A survey on llm-as-a-judge.
\newblock {\em arXiv preprint arXiv:2411.15594}, 2024.

\bibitem{Sakana.AI2024}
Sakana.AI.
\newblock Sakana “ai scientist”.
\newblock {\em Sakana.ai Website}, 2024.

\bibitem{anthropic_claude_code}
Anthropic.
\newblock Claude code: deep coding at terminal velocity.
\newblock Accessed via Anthropic website, 2025.

\bibitem{stojnic2019paperswithcode}
Robert Stojnic and Ross Taylor.
\newblock Papers with code is joining facebook ai.
\newblock Medium article, December 2019.

\bibitem{movassaghi2025articles}
Cameron~S. Movassaghi, Amanda Momenzadeh, and Jesse~G. Meyer.
\newblock From articles to code: on-demand generation of core algorithms from
  scientific publications.
\newblock {\em arXiv preprint arXiv:2507.22324}, 2025.

\bibitem{seo2025paper2code}
Minju Seo, Jinheon Baek, Seongyun Lee, and Sung~Ju Hwang.
\newblock Paper2code: automating code generation from scientific papers in
  machine learning.
\newblock {\em arXiv preprint arXiv:2504.17192}, 2025.

\bibitem{van2025genome}
Camiel~M. van~der Laan, Hill~F. Ip, Marijn Schipper, Jouke-Jan Hottenga, Beate
  St Pourcain, Tetyana Zayats, Ren{\'e} Pool, Eva~M.~L. Krapohl, Isabell
  Brikell, Mar{\'\i}a Soler~Artigas, et~al.
\newblock Genome-wide association meta-analysis of childhood adhd symptoms and
  diagnosis identifies new loci and potential effector genes.
\newblock {\em Nature Genetics}, pages 1--9, 2025.

\end{thebibliography}

\newpage
\section*{Extended Methods}
\subsection*{Details on implementing Paper2Agent}
Paper2Agent converts a research paper and its public codebase into a production-ready MCP server and then exposes that server to an AI agent interface. The process has four stages: (i) codebase identification and extraction, (ii) environment configuration, (iii) tool synthesis and MCP server generation, and (iv) testing, refinement, and deployment, followed by agent connection. We implemented this multi-agent AI system in Claude Code. We design an orchestrator agent that coordinates four sub-agents:
\begin{itemize}
    \item \textbf{Environment-manager}: a specialized agent responsible for creating clean, reproducible environments for research codebases. It analyzes project setup requirements, provisions an isolated workspace, installs all necessary dependencies, and ensures the code runs without conflicts. Standardizing environment setup enables reliable execution and reproducibility across different systems.
    \item \textbf{Tutorial-scanner}: a specialized agent for reviewing the public codebases to identify and organize educational resources. It systematically scans available materials, distinguishes genuine tutorials from other files, and highlights those most useful for reuse. The agent then produces clear summaries and reports, providing a structured view of which resources are worth keeping and which can be set aside.
    \item \textbf{Tutorial-tool-extractor-implementor}: a specialized agent that converts tutorials into reusable tools. It reviews selected tutorials, identifies tasks that generalize beyond the example data, and implements each as a clean, single-purpose function with clear inputs, outputs, and defaults. The agent parameterizes hardcoded values, enforces file-based inputs, saves essential results and figures, and returns a standardized summary of produced artifacts. Its goal is to create a practical function library that reproduces tutorial results on the original data while remaining ready to run on new datasets.
    \item \textbf{Test-verifier-improver}: a specialized agent that creates, runs, and refines tests for tutorial implementations. It uses only the tutorial’s own examples to ensure complete coverage and faithful reproduction of numerical and visualization results. The agent runs in a loop of generating tests, executing them, diagnosing failures, and applying fixes. If functions repeatedly fail, their MCP decorators are removed, and they will not be included in the MCP server. All results and logs are recorded for transparency.
\end{itemize}

Paper2Agent contains six steps:
\begin{enumerate}
    \item \textbf{Locate and download the codebase.} Identify the official repository linked to the paper, clone or download it, and gather associated resources such as supplementary data or configuration files.
    \item \textbf{Environment setup.} Provision a clean, reproducible workspace using an environment manager, pin dependencies, and verify imports so the codebase runs consistently across machines.
    \item \textbf{Tutorial discovery.} Scan the repository to locate useful reference and educational materials and produce an index of candidate tutorials for tooling.
    \item \textbf{Tutorial execution and audit.} Run the selected tutorials end-to-end with their example data, capture inputs, outputs, figures, and runtime constraints, and record any implicit assumptions that must be made explicit.
    \item \textbf{Tool extraction and implementation.} Convert tutorial logic into reusable, single-purpose functions with clear inputs and outputs, parameterize hardcoded values, and save essential artifacts while preserving tutorial fidelity.
    \item \textbf{MCP server assembly.} Integrate the implemented tools, resources, and prompts into a single MCP server with a manifest, versioning, and basic security defaults, ready to be used by an orchestrator or co-scientist agent.
\end{enumerate}

The orchestrator agent invokes sub-agents as needed at different stages of the process. As the Paper2Agent workflow progresses, the results are automatically recorded for each step for traceability and reproducibility. The detailed setup and prompt are available in the \href
{https://github.com/jmiao24/Paper2Agent}{Paper2Agent} GitHub repository.

\subsection*{Generation and analysis of AlphaGenome agent}
We applied the Paper2Agent framework to the AlphaGenome paper to generate an AlphaGenome MCP and connected the MCP with Claude Code to create the AlphaGenome agent. The generated AlphaGenome MCP server is remotely hosted on Hugging Face Spaces (Code availability). To verify reproducibility, the AlphaGenome agent was evaluated using 15 original tutorial-based and 15 novel queries. We prompted the agent with the queries and compared the agent's response with the ground truth answer. The prompt used to query the AlphaGenome agent on interpreting LDL genetic associations is: "\textit{Use AlphaGenome to interpret why chr1:109274968:G>T associates with LDL cholesterol.  Identify the causal genes and assess regulatory effects across modalities in liver. Generate a publication-ready report with figures. My AlphaGenome API key is: <API\_KEY>. Reason step by step.}". The detailed benchmark queries are available in the \href
{https://github.com/jmiao24/Paper2Agent}{Paper2Agent repository}.

\subsection*{Benchmarking the AlphaGenome agent against Claude + Repo and Biomni}
For both the tutorial-based and novel benchmarks described above, we followed these general evaluation steps:
\begin{enumerate}
    \item Generate ground truth answers for each query using manually curated and executed code.
    \item Generate and capture the agent's response to the query, as well as performance metrics like run time and cost.
    \item Manually review and grade the agent's response relative to the ground truth.
    \item Summarize the agent's performance across all queries for the benchmark dataset.
\end{enumerate}
For all agent evaluations, we used claude-sonnet-4-20250514 as the underlying LLM. All evaluations were run locally on a MacBook Air (M2 chip), using model APIs as needed. Each query was run in non-interactive mode from the command line, and all output was captured in JSON format, e.g.:
\begin{verbatim}
bash$ claude --model "claude-sonnet-4-20250514" --print --output-format "json" <prompt>    
\end{verbatim}
Our benchmarking tools and analysis is available in the \href
{https://github.com/jmiao24/Paper2Agent}{Paper2Agent repository}.

\subsubsection*{AlphaGenome agent details}
For the AlphaGenome agent, we used the following context for each query:
\begin{verbatim}
You are an expert in genomics and bioinformatics. 
Please answer the following question using the AlphaGenome MCP tools available to you.

Question: {question}

Please provide a clear, accurate answer based on the AlphaGenome tools. 
If you need to perform calculations or analysis, use the appropriate MCP tools.
You will find the AlphaGenome API key in the .env file in this project.

IMPORTANT: You final response must be a valid JSON object containing exactly two fields:
1. "final_answer": A concise answer containing just the requested value (e.g., a number, 
gene name, or specific result)
2. "reasoning": Your step-by-step reasoning and any calculations you performed

Example response format:
{{
  "final_answer": "GENE_NAME",
  "reasoning": "I used the AlphaGenome variant scoring tool to analyze the variant 
  chr1:1234567:A>C for RNA-seq predictions in Colon - Transverse tissue. 
  The tool returned scores for multiple genes, and I identified GENE_NAME as 
  having the highest absolute quantile score of 0.987."
}}

Please ensure your response is valid JSON and the final_answer field contains 
only the requested value. Do NOT provide any other text or formatting.
\end{verbatim}
\subsubsection*{Claude + Repo agent details}
For the Claude + Repo agent, we used Claude Code with access to a local copy of the \href{https://github.com/google-deepmind/alphagenome}{AlphaGenome repository}. We used the following context for each query:
\begin{verbatim}
You are an expert in genomics and bioinformatics with access to the 
AlphaGenome codebase and API.

Repository: {repo}
File: {path}

Question: {question}

IMPORTANT: You must write and execute Python code using the AlphaGenome library 
to answer this question. Do NOT simply provide answers based on documentation or examples. 
Do NOT provide answers from tutorial notebooks or the executed cells of ipython notebooks. 
You must:

1. **Write Python code** that uses the AlphaGenome API to solve the specific question
2. **Execute the code** and show the results
3. **Extract the exact answer** from the code execution results
4. **Provide the numerical/quantitative answer** that the question is asking for

Key requirements:
- Use `from alphagenome.data import genome` and `from alphagenome.models import dna_client`
- Load the API key from the .env file using `from dotenv import load_dotenv` 
and `load_dotenv()`
- Create the DNA model with `dna_model = dna_client.create(os.getenv('ALPHAGENOME_API_KEY'))`
- Write complete, executable code that addresses the specific question
- Show the code execution results and extract the final answer
- If the question asks for a specific value (like quantile_score, nonzero_mean, etc.), 
provide that exact value

Example approach:
```python
import os
from dotenv import load_dotenv
from alphagenome.data import genome
from alphagenome.models import dna_client

load_dotenv()
dna_model = dna_client.create(os.getenv('ALPHAGENOME_API_KEY'))

# Write code specific to the question here
# Execute the code and show results
# Extract and provide the final answer
```

IMPORTANT: You final response must be a valid JSON object containing exactly two fields:
1. "final_answer": A concise answer containing just the requested value (e.g., a number, 
gene name, or specific result)
2. "reasoning": Your step-by-step reasoning, the Python code you wrote, execution results, 
and any calculations you performed

Example response format:
{{
  "final_answer": "GENE_NAME",
  "reasoning": "I wrote Python code to analyze the variant chr1:1234567:A>C for 
  RNA-seq predictions in Colon - Transverse tissue. The code used the AlphaGenome API to 
  score the variant and identified GENE_NAME as having the highest absolute 
  quantile score of 0.987. Here's the code I executed: [code here] and the results: 
  [results here]."
}}

Please ensure your response is valid JSON and the final_answer field contains only 
the requested value. Do NOT provide any other text or formatting.
\end{verbatim}
\subsubsection*{Biomni agent details}
We utilized the API-based version of Biomni as provided by \href{https://github.com/snap-stanford/Biomni}{the repository}. We used the following context for each query:
\begin{verbatim}
Use AlphaGenome (https://github.com/google-deepmind/alphagenome) to answer the 
following question. You can find the ALPHAGENOME_API_KEY in the .env file for this project.

Question: {question}

IMPORTANT: You final response must be a valid JSON object containing exactly two fields:
1. "final_answer": A concise answer containing just the requested value (e.g., a number, 
gene name, or specific result)
2. "reasoning": Your step-by-step reasoning and any calculations you performed

Example response format:
{{
  "final_answer": "GENE_NAME",
  "reasoning": "I used the AlphaGenome variant scoring tool to analyze the variant 
  chr1:1234567:A>C for RNA-seq predictions in Colon - Transverse tissue. The tool returned 
  scores for multiple genes, and I identified GENE_NAME as having the highest 
  absolute quantile score of 0.987."
}}

Please ensure your response is valid JSON and the final_answer field contains only 
the requested value. Do NOT provide any other text or formatting.
\end{verbatim}

\subsection*{Generation and analysis of TISSUE agent}
We applied the Paper2Agent framework to the TISSUE paper to generate a TISSUE MCP and connect it with the Claude Code to create the TISSUE agent. To assess reproducibility, we compared the TISSUE agent’s outputs against those generated by human researchers using identical Mouse somatosensory cortex ST data. Human researchers performed the analysis based on the tutorial of the TISSUE Github.

\subsection*{Generation and analysis of Scanpy agent}
The Paper2Agent framework was applied to the Scanpy software package to generate a Scanpy agent. This agent was restricted to the preprocessing and clustering workflows within Scanpy, providing a focused and reproducible pipeline for single-cell RNA-seq analysis. The resulting agent was deployed as an MCP server and integrated with Claude Code, enabling natural-language interaction.

To construct the workflow in MCP prompts, we prompted Paper2Agent with: “\textit{Based on the tools you have, construct an MCP prompt to replicate the tutorial in the correct order. Always inspect the data first, and only deviate from the default settings if adhering to them would yield incorrect results.}” This ensured that the generated MCP prompts encoded the standard Scanpy preprocessing and clustering pipeline in a reproducible and interpretable manner.

Reproducibility was evaluated by benchmarking the agent’s outputs against results obtained by human researchers following the official Scanpy reference tutorials. Three publicly available 10x Genomics PBMC single-cell RNA-seq datasets were used as test cases (Data availability). Across datasets, the agent faithfully reproduced all key workflow steps—including gene filtering, normalization, principal component analysis, neighborhood graph construction, and clustering—yielding outputs consistent with human-executed analyses.

\subsection*{Analysis for AI co-scientists linking with AlphaGenome and ADHD GWAS data MCP}

We used the AlphaGenome MCP and the ADHD GWAS data MCP generated by Paper2Agent for this analysis. For the ADHD GWAS data, Paper2Agent automatically extracts relevant information from the main text and supplementary materials of the publication. It parses the supplementary Excel files, cleans the data tables, and converts them into standardized Model Context Protocol (MCP) resources. During this process, descriptive metadata are consolidated into a unified metadata file, ensuring that both data and contextual information are preserved in an agent-readable and reproducible format.

Below is the prompt for scientific hypothesis/question generation.

\begin{verbatim}
You are a expert in scientific hypothesis generation and your job is to propose 10 
actionable  scientific questions that use the tools to explore the data.

You have access to two Model Context Protocol (MCP) servers representing two research 
artifacts (“papers”): MCP A: AlphaGenome and MCP B: ADHD GWAS data. 

Constraints
- Use only resources exposed by AlphaGenome and ADHD GWAS 
data unless instructed otherwise
- No fabricated citations or tool outputs; label missing evidence as "Unknown" 
and propose how to obtain it
- If an MCP tool call fails or returns unexpected data, 
propose alternative approaches or fallback strategies
- You may show reasoning/chain-of-thought separately, 
but the final output must be valid JSON matching the schema below.

Scoring rubric (0–5 each)
- Novelty: originality vs. typical analyses in this domain
- Feasibility: executable with the available MCP tools/resources
- Impact: potential to advance understanding in the target domain
- Validation clarity: measurable success criterion with an 
objective metric
- Resource fit: leverage and synergy of both MCPs’ specific capabilities

For each of the five questions, include
- Title
- One-sentence summary
- Why this combination is promising (3–5 sentences with MCP pointers)
-  Integration sketch: how components connect, key inputs/outputs, and any adapter steps
- Minimal experiment plan: dataset, protocol, primary metric, success threshold, ablations
- Risks and mitigations (explicitly call out data/analysis limitations)
- Resource checklist: which MCP tools/resources you will 
call and in what order (specify server as “AlphaGenome” or “ADHD GWAS data”)
- Individual scores and total score

Output format (strict JSON)

{
  "questions": [
    {
      "title": "",
      "summary_one_sentence": "",
      "rationale": "",
      "integration_sketch": "",
      "experiment_plan": {
        "dataset": "",
        "protocol": "",
        "primary_metric": "",
        "success_threshold": "",
        "ablations": ""
      },
      "risks_and_mitigations": "",
      "resource_checklist": [
        {"server": "AlphaGenome|ADHD GWAS data", "tool_or_resource": "", "purpose": ""}
      ],
      "scores": {
        "novelty": 0,
        "feasibility": 0,
        "impact": 0,
        "validation_clarity": 0,
        "resource_fit": 0,
        "total": 0
      }
    },
    {}, {}, {}, {}
  ],
  "ranking_notes": ""
}
\end{verbatim}

\subsection*{Supplementary Figure}
\begin{figure}[H]
    \centering
    \captionsetup{labelformat=empty}
    \includegraphics[width=0.9\linewidth]{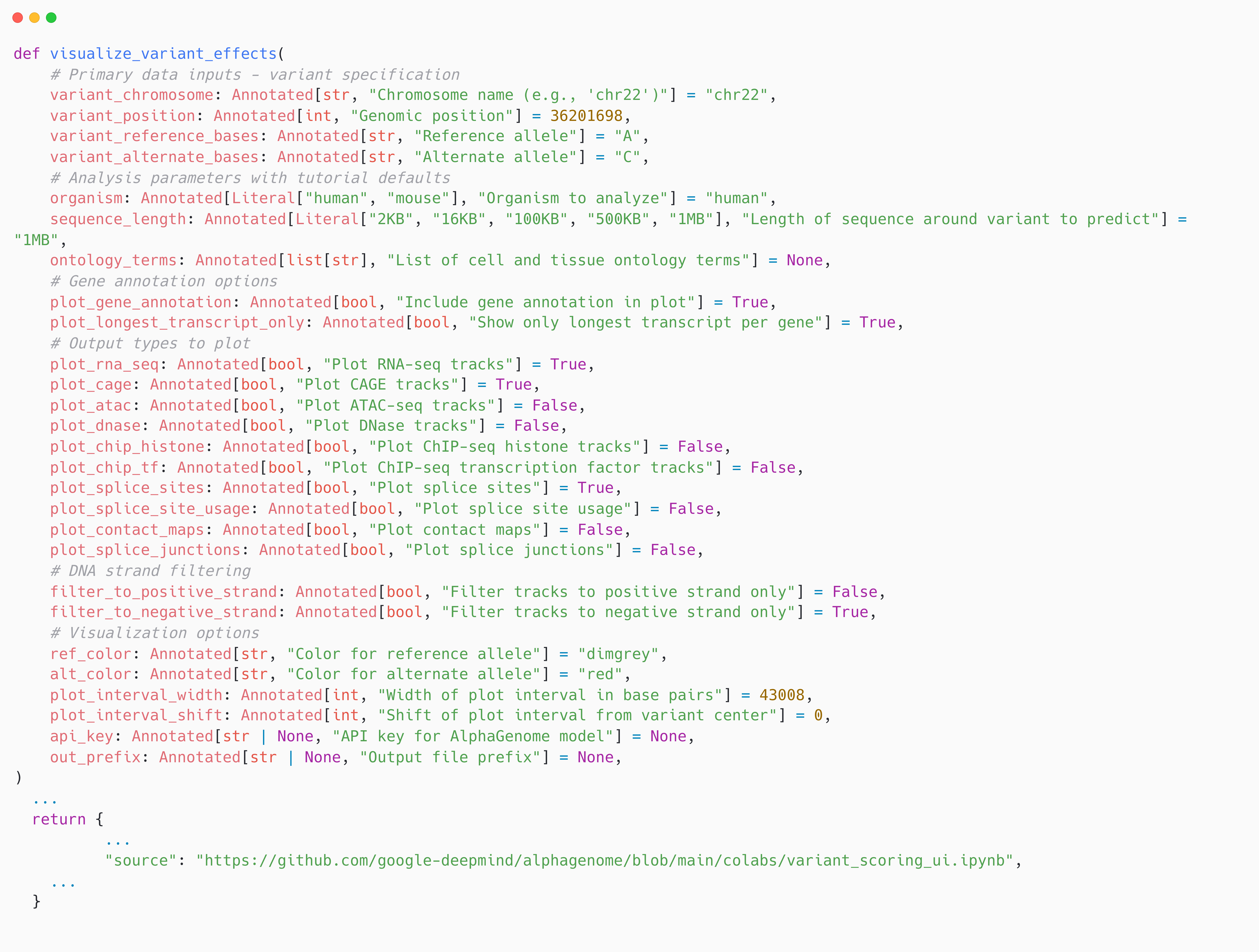}
    \vspace*{-2mm}
    \caption{\textbf{Supplementary Figure 1.} Exposed MCP tools and resources enabling variant scoring and visualization.}
    \label{suppfig:1}
\end{figure}

\end{document}